# Parametrization and Balancing Local and Global Search

Dirk Sudholt[*]

November 8, 2018

## Contents

## 1 Introduction

This chapter is devoted to the parametrization of memetic algorithms and how to find a good balance between global and local search. This is one of the most pressing questions when designing a hybrid algorithm. The idea of hybridization is to combine the advantages of different components. But if one components dominates another one, hybridization may become more hindering than useful and computational effort may be wasted. For the case of memetic algorithms, if the effect of local search is too strong, the algorithm may quickly get stuck in local optima of bad quality. Moreover, the algorithm is likely to rediscover the same local optimum over and over again. Lastly, too much local search quickly leads to a loss of diversity within the population.

The importance of the parametrization of memetic algorithms has already been recognized by Hart [10] in 1994. He posed the following questions, many of which have been reproduced in similar ways in later articles:

- How often should local search be applied?

- On which solutions should local search be used?

- How long should the local search be run?

- How efficient does a local search need to be?

---

[*]The author is with CERCIA, University of Birmingham, Birmingham B15 2TT, UK. This work was done while the author was visiting the International Computer Science Institute, Berkeley, CA, USA.

We will mostly deal with the first and the third question in the sequel. In concrete implementations of memetic algorithms different parameters occur. Related to the first question is a strategy to call local search with a fixed frequency, the *local search frequency*. A similar strategy is to call local search probabilistically, with a fixed *local search probability*. With regard to the third question, often the running time of one local search is capped to a value called *local search depth*. Other mechanisms can have a comparable effect. Ishibuchi, Yoshida, and Murata [13] restricted the neighborhood used for one iteration of local search to some fixed parameter $k$. The size of the neighborhood is also a crucial parameter in variable-neighborhood search algorithms [28]. Paenke, Jin, and Branke [32] used the lifetime of an individual to balance the effect of global- and individual-level adaptation in stochastic environments.

This list of mechanisms for balancing global and local search is by far not complete. While some considerations described in this chapter hold for a large variety of balancing techniques, we will consider the local search frequency and the local search depth as the most typical mechanisms.

We describe the outline of this chapter. In Section 2 we will survey applications and theoretical studies dealing with the parametrization. The effect of local search is discussed and aspects are described that have a strong impact on the optimal balance between global and local search. We also review approaches how to find such an optimal balance. Section 3 deals with the complexity of local search. We will ask how powerful local search is on its own and in which settings a local optimum can be found in polynomial time. For many practically important problems we cannot guarantee that local search always finds a local optimum in polynomial time. Even stronger, there is strong evidence that no algorithm can perform this task in polynomial time. Implications for memetic algorithm design are discussed. Finally, we will present artificial functions in Section 4 and running time analyses demonstrating that the parametrization of memetic algorithms can be extremely hard. This also strengthens the fact that there is no a priori optimal parametrization that works well for every problem. The chapter ends with conclusions in Section 5.

## 2 Balancing Global and Local Search

### 2.1 Early Works and the Effect of Local Search

The early work by Hart [10] and a subsequent extension to combinatorial optimization by Land [20] lead to many conclusions for the design of memetic algorithms. Hart investigated the impact of the local search frequency for the optimization of common test functions in continuous spaces like the Rastrigin function, the Griewank function, and modifications thereof. His experimental results suggest that GAs with large populations are most effective when local search is used infrequently. He also claims that a large local search frequency is needed if the algorithm is not able to identify regions that are likely to contain global optima. As the introduction of elitism increases the degree of exploitation, compared to exploration, less local search is needed when using elitism. Hence, also the type of GA used with local search has a strong impact on performance. Hart also remarks that the use of local search has restricted many applications to use small population sizes because of the increased computational effort. This holds in particular when local search is applied to every individual in the population.

Regarding the selection of individuals for which local search is to be performed, Hart [10] proposes to decrease the local search frequency for each individual by the number of duplicates contained in the population. This works around the problem of having redundant local searches on the same solutions. He generalizes this approach towards reducing the local search frequency with respect to the degree of similarity to other solutions in the population. To this end, a distance metric in genotype space is used. This closely resembles the well-known fitness sharing mechanism for preserving diversity [25]. Land [Section III.C.3 20] proposes several extensions and similar approaches. One is to choose a subset of the population such that the minimum distance between any two selected individuals is maximized. A second strategy is to ensure that every individual in the population is close to an individual that is selected for local search. This way, if the population consists of several clusters, we can hope that all clusters benefit from local search.

Hart [10] also investigated biasing the selection of individuals for local search towards fitter individuals. However, as argued by Land [Section III.C.3 20], this reinforces the dominance of the already fit individuals and hence leads to a rapid loss of diversity. In addition, good solutions are likely to be close to local optima, hence they will have the least benefit of applying more local search to them. Also, for solutions that are close to local optima improvements may be hard to find, which renders the local search less efficient.

Related to the last remark is the question how easy improvements can be achieved for specific solutions. Land [Section III.C.3 20] introduced the notion of a "local search potential" as a measure for the expected gain in fitness in relation to the computational effort. The local search potential can be estimated by performing few steps of local search, a so-called "local search sniff" and recording both the gain and the effort throughout the sniffing period. The average gain per unit of effort is then used as an estimation for its future effectiveness. The drawbacks of this approach is that these sniffs might use a fair amount of computational effort to yield reliable estimations. Moreover, there are no guarantees that the progress in early steps of local search will be an accurate prediction of future progress.

The use of local search is not restricted to evolutionary algorithms. Memetic approaches have also been used for various other paradigms such as estimation-of-distribution algorithms [1] or Ant Colony Optimization [4, 22]. The effect of local search can be quite different in other paradigms. In a recent study Neumann, Sudholt, and Witt [30] argued that the use of local search in ant colony optimization (ACO) can change the behavior of the algorithm drastically. Without local search, the sampling distribution for new solutions given by artificial pheromones usually follows the best-so-far solution. This enables the algorithm to follow paths and ridges in the search space. When introducing local search with a large local search depth, however, a newly discovered local optimum might be far away from the "center of gravity" of the sampling distribution. In ACO algorithms using the best-so-far rule (i.e. always rewarding the current best solution found so far), the pheromones are then directly adapted towards the new local optimum. Instead of following the path taken by local search to arrive at this local optimum, the direct adaptation of pheromones can make the algorithm sample solutions from a totally different area of the search space. Neumann et al. [30] demonstrated for a constructed function where this effect may mislead the search and turn a polynomial optimization time into an exponential one, with high probability. However, they also proved for a slightly different function that this behavior can also prevent the algorithm from getting stuck in a local optima. Local search can then also help to reduce an exponential optimization time to a polynomial one.

## 2.2 Aspects that Determine the Optimal Balance

The optimal balance between global and local search clearly depends on the optimization problem at hand and the memetic algorithm applied to it. The latter not only includes the choice of the operators employed and issues of representation, but also various other parameters of the algorithm such as the population size, selection pressure, and the mutation rate. Even among the mentioned aspects and for plain evolutionary algorithms there is strong evidence that the precise choice of parameters can have a tremendous effect on performance. Theoretical studies were performed, e.g., by Storch [38] and Witt [46] for the choice of the parent population size, Jansen, De Jong, and Wegener [15] for the choice of the offspring population size, Jansen and Wegener [14] for the choice of the mutation rate, and Lehre and Yao [21] for the ratio of the selection pressure in ranking selection and the mutation rate.

We therefore cannot expect to obtain design guidelines that do not depend on all the mentioned aspects and nevertheless always lead to good results. The existence of such guidelines is excluded by the well-known *no free lunch theorems* [12, 24]. These results state that when averaging over a class of problems that is closed under permutation, all algorithms (this includes all parametrizations for one specific algorithm) have equal average performance. It is, however, also clear that the setting of the no free lunch theorems is much too general to be of any relevance. The vast majority of functions considered are of no interest for optimization as they have exponential-size representations [6]. In Section 4 we will present much stronger results for one particular memetic algorithm. The considered functions do have polynomial-size representations and exhibit superpolynomial or exponential performance gaps for even small changes of the parametrization. This shows that for the considered algorithm there is no polynomial relation between optimal and non-optimal parameter values.

So, the parameters and design aspects of a memetic algorithm should not be viewed in isolation. The strongest dependency is probably the one between the local search depth and the local search frequency.

|                       | less local search         | more local search                       |
|-----------------------|---------------------------|------------------------------------------|
| exploration by GA     | weak exploration          | strong exploration                       |
| mutation strength     | small mutations           | large perturbations                      |
| pivoting rule         | steepest ascent/descent   | first improvement                        |
| neighborhood size     | large neighborhood        | small neighborhood/reduction techniques  |
| implementation of LS  | expensive recalculations  | incremental fitness evaluations          |
| objectives            | multi-objective problem   | single-objective problem                 |

Table 1: Overview on aspects that affect the optimal amount of local search.

Choosing one parameter value with disregard to the other one often does not make much sense. For instance, Ishibuchi et al. [13] discovered that in applications to the multi-objective permutation flowshop scheduling problem the optimal number $k$ of neighbors visited in one iteration of local search was strongly negatively correlated with the local search frequency $p_{\text{LS}}$. The best performance was obtained when the product $k \cdot p_{\text{LS}}$ was within a range of 1 to 10.

Also, the balance of exploration and exploitation is important. In iterated local search algorithms [23] local search is typically used in every iteration and performed until a local optimum is found. So, local search is used to its utmost extend. On the other hand, iterated local search algorithms tend to use strong perturbations, i.e., large mutations before applying local search. In this setting, a powerful explorative operator balances out a powerful exploitative operator. When the underlying evolutionary component of a memetic algorithm is more similar to a classical genetic algorithm, that is, if more emphasis is put on exploration by populations and the use of recombination and mutation, less local search should be used in order not to disrupt exploration.

The optimal balance between global and local search also depends on design and implementation issues. In some applications, local search is computationally expensive. This holds, for example, in the case of large or computationally expensive neighborhoods like the Lin-Kernighan neighborhood or pivoting rules such as steepest descent/ascent, where the whole neighborhood must be searched. Using pivoting rules such as first improvement or neighborhood reduction techniques can speed up the local search significantly and thus shift the "optimal" amount of local search.

In several applications it is possible to perform incremental fitness evaluations during local search. If the fitness can be efficiently updated in case only few components (bits, objects, edges, ...) are modified in an iteration of local search, local search tends to be much faster than the genetic component of the algorithm. One example is the TSP where the cost of a 2-Exchange operation can be computed by only looking at the 4 edges involved, see, e.g. Ishibuchi et al. [13], Merz [26]. In fact, Jaszkiewicz [16] reported in a study on a multi-objective TSP problem that local search was able to perform 300 times more function evaluations per second than a multi-objective genetic algorithm. Also neighborhood reduction techniques turned out to be very useful for speeding up local search [26].

On the other hand, Ishibuchi et al. [13] argued that for flowshop scheduling recomputing the fitness after a local change of a schedule cannot be done much faster than computing the fitness from scratch. This is because even local changes may imply that the completion times for almost all jobs have to be recalculated. The execution time for one iteration of local search is thus a very important issue.

When considering multi-objective problems, it is important to maintain diversity in the population. Sindhya, Deb, and Miettinen [35] used a local search that optimizes an achievement scalarizing function. The local search helps with the convergence to the Pareto front, but it is also likely to create extreme points on the Pareto front. To this end, the authors used a dynamic schedule for choosing the local search probability. The local search probability linearly increases from 0 to the inverse population size and then drops to 0 again. The number of generations for one such cycle is proportional to the population size.

Concluding, there are many aspects that determine the optimal balance between global and local search. Many different parameter settings have been proposed, some of which are due to dynamic or adaptive schedules. Table 2.2 summarizes the above-mentioned aspects. In the following, we will describe approaches how such an optimal balance can be found.

## 2.3 How to Find an Optimal Balance

Several approaches have been proposed how to find a good parametrization for memetic algorithms. There are general approaches for finding good parameter settings that are not tailored towards memetic algorithms and hence are somewhat beyond the scope of this chapter. We brief mention one such approach called sequential parameter optimization (SPO) introduced by Bartz-Beielstein, Lasarczyk, and Preuß [3]. SPO aims at finding the best parametrization by combining classical and modern statistical techniques. It can be seen as a search heuristic trying to optimize the performance of non-deterministic algorithms. SPO iteratively applies the following three steps. First, an experimental analysis of an algorithm with a given parametrization is performed. Then, the performance of the algorithm (including its parametrization) is estimated by means of a stochastic process model. In a third step, additional parameter settings in the parameter space are determined in a systematic way. For further details, we refer to Bartz-Beielstein [2], Bartz-Beielstein, Lasarczyk, and Preuß [3].

Goldberg and Voessner [9] and Sinha, Chen, and Goldberg [36] presented a system-level theoretical framework for optimizing global-local hybrids. Two different optimization goals are considered: maximizing the probability of reaching a solution within a given accuracy and minimizing the time needed to do so. The authors considered the impact of the local search depth for a hybrid that uses random search as global component. They presented formulas for the determining the optimal local search depth for the mentioned optimization goals. The formulas, however, are based on some simplifying assumptions and they do require knowledge on the structure of the problem that is usually not available in practice. The probabilities of reaching specific basins of attraction in one step of the global searcher have to be known as well as the average time local search takes to local optimality for each basin.

Another well-studied approach is to include domain knowledge into the design of memetic algorithms [26, 43]. This knowledge can be gained by analyzing the fitness landscape of the problem (instance) at hand. One useful measure for the ruggedness of a fitness landscape is the *correlation length*. It is, in turn, based on the *random walk correlation function* $r(s)$, also known as *autocorrelation*. The function $r(s)$ specifies the correlation between two points of a random walk that are $s$ time steps away. The random walk chooses the next point uniformly from a fixed neighborhood. Different neighborhoods may thus lead to different correlations. If the correlation is high, the correlation length is large and the fitness landscape is smooth. If the correlation is low, the correlation length is small and the fitness landscape is rugged. It has been observed that large correlation lengths lead to a large number of iterations until local search finds a local optimum. On the other hand, a small correlation length often means that local search may quickly get stuck in bad local optima [26]. Fitness landscape analysis can help to choose the right neighborhood and a suitable parametrization for the local search.

Last but not least, adaptive techniques may help to find a good parametrization. Memetic algorithms using many different local searchers are known as *multimeme algorithms* [29]; each local search operator is called a "meme." The choice of memes can be made adaptively or even self-adaptively, see the survey by Ong, Lim, Zhu, and Wong [31]. Also coevolutionary systems have been developed that coevolve a local searcher alongside the evolution of solutions [18, 37]. As a discussion of adaptive systems is beyond the scope of this chapter, we refer to Chapter II.7 for further details.

## 3 Time Complexity of Local Search

In order to fully understand the capabilities of local search, it is indispensable to know its limitations. In this section we describe theoretical results on the time complexity of local search and discuss implications on memetic algorithm design. We will look at local search in isolation and ask how long it takes until one call of local search finds a local optimum. From the perspective of memetic algorithms, we ask how efficient the local search component is in computing a local optimum from its basin of attraction. If local search cannot find local optima efficiently, a memetic algorithm will most likely show poor performance, even if the global component can locate the basin of attraction of the global optimum efficiently. We will also review a theory of intractability that applies to many important problems and memetic algorithms used in practice. It can be proven that under certain complexity-theoretic assumptions and in the worst case local optima cannot be computed in polynomial time by any means, even for more sophisticated algorithms than local search. It is not the case that local search is too simple to locate local optima efficiently. Instead, the mentioned problems are so difficult that computing local optima is hard for any (arbitrarily sophisticated) search strategy.

The following presentation is based in parts on Michiels, Aarts, and Korst [27, Chapter 6]. Define a *local search problem* as a combination of a combinatorial optimization problem, a neighborhood function mapping a solution to a subset of the search space, and an indication whether the problem is a maximization or a minimization problem. The goal of a local search problem is to compute a local optimum with respect to the goal of the optimization. Note that the neighborhood is an integral part of the problem. Using a different neighborhood function leads to a different local search problem.

The main question is how many iterations local search will need in order to find a local optimum. It is helpful to use the following perspective. Define the state graph of a problem as a directed graph where the set of vertices corresponds to the search space. The state graph includes an edge $(x, y)$ if and only if $y$ is a neighbor of $x$ and $y$ is strictly better than $x$. A local optimum thus corresponds to a sink, i.e., a vertex with no outgoing edges. The number of iterations needed to find a local optimum corresponds to the length of the path from the starting point to a sink. The precise choice of an outgoing edge is determined by the pivoting rule.

## 3.1 Polynomial and Exponential Times to Local Optimality

In many applications, local search finds an optimum in polynomial time. Assume the neighborhood is searchable in polynomial time and the number of function values is polynomially bounded. Then clearly all paths in the state graph only have polynomial length and local search will finish in polynomial time. Problems with only a polynomial number of function values include the NP-hard Minimum Graph Coloring problem if the number of colors used is taken as fitness function and the NP-hard MAXSAT problem, when one uses the number of satisfied clauses as objective function. Another NP-hard problem with this property is the graph partitioning problem. The fitness corresponds to the number of cut edges, which ranges from 0 to $n^2/4$, $n$ being the number of vertices. Also weighted problems might show this property, for instance in special cases where the weights are integral, positive, and polynomially bounded. Land [20, Section III.A.1] gives a formal proof for a class of weighted graph partitioning problems and a weighted TSP.

Lin-Kernighan-type or variable-depth-type of local searches perform a chained sequence of local moves and fix solution components (edges, bits, vertices, ...) that have been changed until the end of local search. Hence, these local searches also trivially stop after polynomially many steps (see [42] for an analysis of memetic algorithms with variable-depth search). The effect is similar as for local searches with a maximum local search depth; local search stops in polynomial time without guarantee of having found a local optimum.

When the number of function values is superpolynomial, it might still be that all paths in the state graph have only polynomial length. But for some problems one can actually prove that in settings with exponentially many function values exponentially long paths exist. Englert, Röglin, and Vöcking [7] constructed an instance for the Euclidean TSP where the state graph for the 2-Opt algorithm has exponential length. Hence, in the worst case—with respect to the choice of the starting point and the pivoting rule—local search takes exponential time.

Similar results also hold for pseudo-Boolean optimization. Horn, Goldberg, and Deb [11] presented so-called *long path problems* which contain a fitness-increasing path in the state graph under the Hamming neighborhood (two solutions are neighbored if they only differ in exactly one bit). The length of the path is of order $\Theta(2^{n/2})$ if $n$ is the number of bits. In addition, for every point $x$ on the path every Hamming neighbor $y$ of $x$ has strictly lower fitness than $x$, unless $y$ is itself a point on the path. In other words, the next successor on the path is the only neighbor with a better fitness. This property ensures that a local search using the Hamming neighborhood cannot leave the path and thus is forced to climb to its very end. This holds regardless of the pivoting rule as the pivoting rule cannot make any choices. All points not belonging to the path give hints to reach the start of the path, hence also on average over all starting points local search needs exponential time.

Note, however, that flipping 2 bits at a time or using a stochastic neighborhood such as standard bit mutations suffices to reach the end of the path efficiently by taking shortcuts. Rudolph [34] proved an upper bound of $O(n^3)$ for the expected optimization time of the simple algorithm (1+1) EA whose mutation operator flips each bit independently with probability $1/n$. He also formally defined a more robust generalization to long $k$-paths where at least $k$ bits have to flip in order to take a shortcut. The parameter $k$ can be chosen such that the length of the path is still exponential (say, of order $2^{\sqrt{n}}$) and the probability of taking a shortcut by standard bit mutations is still exponentially small. This yields an example where also using larger neighborhoods that can flip up to $k-1$ bits at a time need exponential time for suitable initializations. Also the stochastic neighborhood used by the (1+1) EA does not avoid exponential expected optimization times for suitable values of $k$, as proven by Droste, Jansen, and Wegener [5].

## 3.2 Intractability of Local Search Problems

NP-completeness theory is a well-known and powerful tool to prove that many important optimization problems are intractable, in a sense that no polynomial-time algorithm for the problem can exist, assuming P ≠ NP. There is a similar theory for local search problems that can be used to characterize local search problems where under reasonable assumptions no polynomial-time algorithm exists for finding local optima. This includes arbitrary algorithms that need not have much in common with local search algorithms. The foundation for this theory was laid by Johnson, Papadimitriou, and Yannakakis [17]. We give an informal introduction into this theory and refer the reader to Yannakakis [47] and Michiels et al. [27, Chapter 6] for complete formal definitions. For this subsection we assume that the reader has basic knowledge on NP-completeness and refer to classical text books for further reading [8, 33, 44]. A brief treatment of NP-completeness is also given in Michiels et al. [27, Appendix B].

The complexity class we will focus on is called PLS for "polynomial-time searchable." A local search problem $\Pi$ is in PLS if there exist two polynomial-time algorithms with the following properties. One algorithm can be seen as an initialization operator. It simply computes some initial solution for $\Pi$ in polynomial time. The second polynomial-time algorithm, given a solution $s$, either computes a better neighbor of $s$ or reports that $s$ is a local optimum. If a problem is in PLS, this means that there is a local search algorithm such that the initialization and each iteration of local search can be executed in polynomial time. This is not to be confused with the question how many iterations are needed in order to find a local optimum.

Similar to reductions in NP-completeness theory, there is the concept of a reduction between PLS-problems: we can relate the difficulties of two problems $\Pi_1, \Pi_2$ in PLS as follows. Denote a PLS-reduction from $\Pi_1$ to $\Pi_2$ by $\Pi_1 \leq_{\text{PLS}} \Pi_2$. A PLS-reduction demands a polynomial-time algorithm that maps a problem instance of $\Pi_1$ to an instance of $\Pi_2$ and a polynomial-time algorithm that maps a solution for $\Pi_2$ back to a solution for $\Pi_1$. In the latter mapping, we require that if the solution $s_2$ for $\Pi_2$ is a local optimum for $\Pi_2$ and $s_2$ is mapped to a solution $s_1$ for $\Pi_1$, then $s_1$ must be a local optimum for $\Pi_1$. Hence, if we want to solve $\Pi_1$, we can use the first algorithm to transform the instance for problem $\Pi_1$ into an instance of $\Pi_2$, then solve problem $\Pi_2$ to local optimality, and finally map the local optimum back to a local optimum for $\Pi_1$ using the second algorithm.

If $\Pi_1 \leq_{\text{PLS}} \Pi_2$ then we can conclude that $\Pi_2$ is "at least as hard" as $\Pi_1$. This means that if $\Pi_1$ cannot be solved in polynomial time, then $\Pi_2$ cannot be solved in polynomial time either. But if $\Pi_2$ is polynomial-time solvable, then $\Pi_1$ also is. This concept leads to the notion of PLS-completeness: a problem $\Pi$ is PLS-complete if *every* problem in PLS can be PLS-reduced to it; in other words, $\Pi$ is PLS-complete if it is at least as hard as every other problem in PLS. PLS-complete problems thus constitute the hardest problems in PLS. If it could be shown for one PLS-complete problem that a local optimum can always be found within polynomial time, then all problems in PLS would be solvable in polynomial time. Speaking in terms of complexity classes, we would then have P = PLS. However, as no polynomial-time algorithm has been found for *any* PLS-complete problem, it is widely believed that P ≠ PLS.

**Theorem 1.** *If $P \neq PLS$, there exists no algorithm that always computes a local optimum for a PLS-complete local search problem in polynomial time.*

This result not only states that local search probably cannot find local optima for PLS-complete problems. It also says that no other, arbitrarily sophisticated algorithm can do better.

The theory of PLS-completeness has concrete implications as many well-known local search problems have been proven to be PLS-complete. We list some examples and refer to Michiels et al. [27, Appendix C] for a more detailed list.

**Theorem 2.** *The following local search problems are PLS-complete.*

- *Pseudo-Boolean optimization: maximize or minimize a function $\{0,1\}^n \to \mathbb{R}$ using the Hamming neighborhood*

- *MAX-2-SAT for the Hamming neighborhood as well as the Kernighan-Lin neighborhood*

- *MAXCUT for the Hamming neighborhood as well as the Kernighan-Lin neighborhood*

- *Metric TSP for the k-Exchange neighborhood as well as (a slightly modified variant of) the Lin-Kernighan neighborhood.*

So, there are PLS-completeness results for neighborhoods used by common local search algorithms. Memetic algorithms usually combine different neighborhoods for genetic operators and local searchers. Multimeme algorithms or variable-neighborhood search even use several neighborhoods for local search. Does PLS-completeness also hold in these settings?

The answer is yes. Recall that a PLS-reduction $\Pi_1 \leq_{\text{PLS}} \Pi_2$ demands that all local optima in $\Pi_2$ must be mapped to local optima in $\Pi_1$. In order to prove that a problem $\Pi_2$ is PLS-complete, it suffices to show that $\Pi_1 \leq_{\text{PLS}} \Pi_2$ for a PLS-complete local search problem $\Pi_1$. Assume that $\Pi_1$ is PLS-complete and consider the situation where $\Pi_1$ and $\Pi_2$ are based on the same combinatorial problem. Further assume that $\Pi_2$ uses a "larger" neighborhood in the following sense: if $x$ and $y$ are neighbored in $\Pi_1$ then they are also neighbored in $\Pi_2$. For example, $\Pi_1$ might be the TSP with a 2-Exchange neighborhood and $\Pi_2$ might be the TSP with a neighborhood of all 2-Exchange and 3-Exchange moves. Now, if $x$ is a local optimum in $\Pi_2$ then it is also a local optimum in $\Pi_1$ (it might even have less neighbors to compete with). Hence, using the identity function for mapping local optima in $\Pi_2$ back to $\Pi_1$ establishes a PLS-reduction $\Pi_1 \leq_{\text{PLS}} \Pi_2$ and proves PLS-completeness for $\Pi_2$. Note that the term "neighborhood" can be used in a broad sense. In the above example, $\Pi_2$ might use different kinds of operators. For instance, instead of containing 2-Exchange and 3-Exchange moves, the neighborhood of $\Pi_2$ could contain 2-Exchange moves and Lin-Kernighan moves. Note, however, that the enlarged neighborhood must still be searchable in polynomial time as otherwise $\Pi_2$ would not be contained in PLS.

It is also possible to incorporate populations as described by Krasnogor and Smith [19]. A local search problem $\Pi_1$ whose state space reflects a single solution can be mapped to a local search problem $\Pi_2$ whose state space reflects all possible populations. The function value for $\Pi_2$ can be defined as the $\Pi_1$-value for the best individual in the population. The neighborhood function for $\Pi_2$ would contain all possible transitions to other populations using the neighborhood function of $\Pi_1$. As long as this neighborhood is searchable in polynomial time, a PLS-reduction $\Pi_1 \leq_{\text{PLS}} \Pi_2$ can simply map the best individual from the population of the problem $\Pi_2$ to $\Pi_1$. If the population cannot be improved by any operation in $\Pi_2$, then the best individual cannot be improved in $\Pi_1$. Hence, a locally optimal population for $\Pi_2$ implies a locally optimal individual for $\Pi_1$. With this PLS-reduction, we have shown that the population-enhanced problem $\Pi_2$ is PLS-complete as well.

The conclusion from these observations is the following: if we know that a local search problem $\Pi_1$ is PLS-complete, then all algorithms that result from $\Pi_1$ by extending the algorithm to populations, enlarging neighborhoods, or adding new operators are, in turn, PLS-complete. This holds under the condition that all considered neighborhoods are searchable in polynomial time. Krasnogor and Smith [19] formalize PLS-completeness results for memetic algorithms on the TSP that use the 2-Opt operator. Quoting from their work, "the addition of a population to the evolutionary heuristic does not improve the worst-case behavior beyond that of local search."

For the sake of completeness, we also mention that there is a stronger notion of PLS-completeness, called *tight PLS-completeness*. For tightly PLS-complete problems there can exist paths in the state graph of exponential length. This implies that local search needs exponential time in the worst case. This holds even regardless of the pivoting rule. Actually, all problems mentioned in Theorem 2 are tightly PLS-complete. To prove tight PLS-completeness, so called *tight PLS-reductions* are needed that additionally preserve the length of paths in the state graph, up to polynomial factors. Tight PLS-completeness is, however, not robust w.r.t. extensions of the neighborhood as larger neighborhoods might add shortcuts in the state graph.

How can we deal with PLS-complete problems? Recall that PLS-completeness only focusses on the worst-case behavior. Even if the worst case is hard, the average-case performance or the performance when starting with "typical" starting points generated by the global component might be much better. In fact, problem instances constructed to reveal exponential-length paths in the state space are mostly contrived and very dissimilar to problem instances encountered in practice. Furthermore, even if there is an intractability result for a general problem, it might be that one is actually solving an easier special case of the general problem. While the TSP using common neighborhoods is PLS-complete for general edge weights, local search trivially succeeds in polynomial time if the edge weights are positive and polynomially bounded integers. Though the general problem is (tightly) PLS-complete, the weight-restricted TSP is not.

## 4  Functions with Superpolynomial Performance Gaps

From general hardness results that hold for classes of algorithms under certain assumptions, we now move on the more concrete results for specific memetic algorithms. We will present results that prove the non-existence of a priori guidelines for the parametrization of the investigated memetic algorithms. For both the local search depth and the local search frequency there are functions where only specific parameter values can guarantee an effective running time behavior. With only small variations of the parameters, the typical running time experiences a phase transition from polynomial to superpolynomial or even exponential running times. The "optimal" parameter values for these functions can be chosen almost arbitrarily. This implies that for almost each fixed parametrization (whose value may depend on the problem size) there is a function for which this parameter is far from being optimal. This section is based on Sudholt [41]. Preliminary results were published in Sudholt [39, 40].

The non-existence of an all-purpose optimal parameter value is not surprising in the light of the no free lunch theorems [12, 24], but our statements are much stronger. For instance, they prove that the running times of "good" and "bad" parameter values are not polynomially related. Also, the no free lunch theorems only yield a mere existence proof and do not give any hints how separating functions might look like.

The downside of this approach is that these strong statements can only be obtained by fixing a memetic algorithm that is simple enough to be handled analytically. In particular, the algorithm does not use crossover. The algorithm is called $(\mu+\lambda)$ EA. It uses a fixed maximum local search depth denoted by $\delta$ and calls local search with a fixed frequency, every $\tau$ iterations. The local search used iteratively searches for neighbors with strictly larger fitness and stops if no such point exists or the maximum local search depth of $\delta$ iterations has been hit. It may be implemented using an arbitrary pivoting rule. The $(\mu+\lambda)$ MA operates

---

**Algorithm 1** Local search($y$)

**for** $\delta$ iterations **do**
    **if** there is a $z \in N(()y)$ with $f(z) > f(y)$ **then** $y := z$
    **else** stop and **return** $y$.
**return** $y$.

---

with a population of size $\mu$ and creates $\lambda$ offspring in each generation. This is done by choosing randomly a parent, then mutating it, and, every $\tau$ generations, additionally applying local search to the result of the mutation. The population for the next generation is selected among the best parents and offspring.

---

**Algorithm 2** ($\mu$+$\lambda$) Memetic Algorithm

---

Let $t := 0$.
Initialize $P_0$ with $\mu$ individuals chosen uniformly at random.
**repeat**
    $P'_t := \emptyset$.
    Do $\lambda$ times:
        Choose $x \in P_t$ uniformly at random.
        Create $y$ by flipping each bit in $x$ independently with prob. $p_m$.
        **if** $t \bmod \tau = 0$ **then** $y := \text{local search}(y)$.
        $P'_t := P'_t \cup \{y\}$.
    Create $P_{t+1}$ by selecting the best $\mu$ individuals from $P_t \cup P'_t$. (Break ties in favor of $P'_t$.)
    $t := t + 1$.

---

## 4.1 Functions where the Local Search Depth is Essential

Now we describe how to construct a function $f_D$ parametrized by an "ideal" value $D$ for the local search depth, such that the following holds. Formal definitions can be found in Sudholt [41]. If the local search depth is chosen as $\delta = D$, then the ($\mu$+$\lambda$) EA optimizes $f_D$ efficiently. However, if the local search depth is only a little bit away from this ideal value, formally $|\delta - D| \geq \log^3 n$, then the ($\mu$+$\lambda$) EA needs superpolynomial time, with high probability. The precise result reads as follows.

**Theorem 3.** *Let $D \geq 2\log^3 n$, $\lambda = O(\mu)$, and $\mu, \delta, \tau \in \text{poly}(n)$. Initialize the ($\mu$+$\lambda$) MA with $\mu$ copies of the first point on the path, then the following holds with high probability:*

– *if $\delta = D$, the ($\mu$+$\lambda$) MA optimizes $f_D$ in polynomial time*

– *if $|\delta - D| \geq \log^3 n$, the ($\mu$+$\lambda$) MA needs superpolynomial time on $f_D$.*

We only remark without giving a formal proof that the function can be adapted such that in the second case the stronger assumption $|\delta - D| \geq n^\varepsilon$ for some constant $\varepsilon > 0$ leads to exponential optimization times.

In the following, we describe the construction of the function $f_D$ and the main proof ideas. The construction is based on the long $k$-paths already mentioned in Section 3. On this path it is very unlikely that mutation can find a shortcut as at least $k = \Omega(\sqrt{n})$ bits would have to flip simultaneously in one mutation. For simplicity, we assume that the algorithm starts with the whole population at the start of the path and only mention that the construction can be adapted for random initialization. All points that are neither on the path nor global optima are assigned a very low fitness, so that the algorithm only searches on the path. In fact, the mentioned points all receive the same low fitness value, so that local search stops immediately if called from a point that is surrounded by low-fitness individuals. In some sense, we have thus transformed an $n$-dimensional problem into a one-dimensional problem. The path points are assigned fitness values in the following way. The basic idea is that a global optimum can only be found with good probability if local search stops at specific points on the long $k$-path.

The path can be divided into sections on which the fitness is strictly increasing on the path. Each section ends with a local optimum. A sketch of the function is given in Figure 1, reproduced from Sudholt [41]. The absolute fitness values at the start of these sections is set so low that a section can only be climbed by local search, given that a preceding mutation creates a suitable starting point. For each section, a set of global optima is placed close to the path in a way that local search cannot locate a global optimum when climbing the section. (This is where we have to think of an $n$-dimensional problem again as this cannot be properly drawn in a one-dimensional picture.) However, if the local search depth is set in such a way that local search stops close to the global optima, then there is a good chance of jumping to a global optimum the next time this individual is selected for mutation.

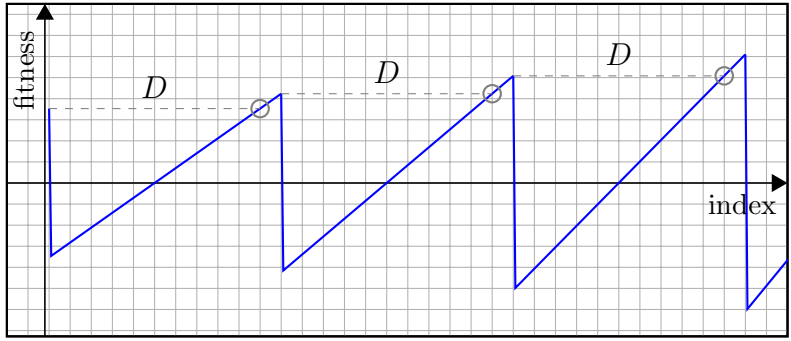

Figure 1: Sketch of the function $f_D$. The $x$-axis shows the index on the long $k$-path. The $y$-axis shows the fitness. The thick solid line shows the fitness of the points on the long $k$-path. Encircled path points are close to a target region with respect to Hamming distance. The long $k$-path can be separated into $n$ subsequent sections with increasing fitness, each one ending with a local optimum. For the sake of clarity, only the first three out of $n$ sections are shown.

Now, the main ideas of the proof are as follows. If the local search depth is smaller than $D - \log^3 n$, local search typically stops with a search point that is inferior to all points in the population. The new offspring is then immediately rejected by selection. The only way to avoid this is to make a large jump by mutation that flips at least $\log^3 n$ bits simultaneously. The probability for this event is superpolynomially small and the expected waiting time until this happens is superpolynomial. This establishes a superpolynomial lower bound in the case $\delta \leq D - \log^3 n$.

In case the local search depth attains the "ideal" value $\delta = D$, there is a constant probability that local search stops close to a set of global optima and a mutation flipping two bits creates a global optimum next time this individual is selected for mutation. Note that the algorithm only needs to be successful on one section in order to find a global optimum. With high probability this happens at least once within $n$ trials and the algorithm succeeds in polynomial time.

If the local search depth is too high, i.e., $\delta \geq D + \log^3 n$, every time local search climbs a section it runs past the set of global optima and ends with the next local optimum. This holds since each section has length $D + \log^3 n$. From there, a global optimum can only be reached by a large mutation or if the population is able to approach the target set by moving downhill on the section from the local optimum. Note that a new offspring might survive even if it is worse than its parent in case the population contains individuals that are still worse than the offspring. However, using family-tree techniques [45], one can prove that with high probability the population is quickly taken over by the best individuals in the population before getting downhill.

## 4.2 Functions where the Local Search Frequency is Essential

Also the choice of the local search frequency can have a tremendous impact on the performance of the $(\mu+\lambda)$ MA. As the analysis presented in Sudholt [41] is quite involved, the results are limited to the $(1+1)$ MA where $\mu = \lambda = 1$. Two functions called $\text{Race}^{\text{con}}$ and $\text{Race}^{\text{uncon}}$ are defined according to given values for $n, \delta$, and $\tau$. For formal definitions we again refer to Sudholt [41]. The $(1+1)$ MA is efficient on $\text{Race}^{\text{con}}$, but inefficient on $\text{Race}^{\text{uncon}}$. Now, if the local search frequency is halved, the $(1+1)$ MA suddenly becomes inefficient on $\text{Race}^{\text{con}}$, but efficient on $\text{Race}^{\text{uncon}}$.

The functions $\text{Race}^{\text{con}}$ and $\text{Race}^{\text{uncon}}$, which we call *race functions*, are constructed in similar ways, so we describe them both at once. First of all, we partition all bit strings into their left and right halves, which form two subspaces $\{0,1\}^{n/2}$ within the original space $\{0,1\}^n$ for even $n$. Each subspace contains a part of a long path. Except for special cases, the fitness is the (weighted) sum of the positions on the two paths. This way, climbing either path is rewarded and the (1+1) MA is encouraged to climb both paths in parallel.

The difference between the two paths in the left and right halves of the bit string is that they are adapted to the two neighborhoods used by mutation and local search, respectively. In the left half, we have a connected path of predefined length. The right half contains a path where only every third point of the long $k$-path is present. Instead of a connected path, we have a sequence of isolated peaks where the closest peaks have Hamming distance 3. As the peaks form a path of peaks, we speak of an *unconnected path*. While the unconnected path cannot be climbed by local search, mutation can jump from peak to peak as a mutation of 3 specific bits has probability at least $1/(en^3)$. Concluding, local search is well suited to climb the connected path while mutation is well suited to climb the unconnected path.

Now, the main idea is as follows. Choosing appropriate lengths for the two paths, if the local search frequency is high, we expect the (1+1) MA to optimize the connected path prior to the unconnected path. Contrarily, if the local search frequency is low, the (1+1) MA is likely to optimize the unconnected path prior to the connected one. Which path is optimized first can make a large performance difference. In the special cases where the end of any path is reached, we define separate fitness values for Race$^{\text{con}}$ and Race$^{\text{uncon}}$. For Race$^{\text{con}}$, if the connected path is optimized first (i.e., wins the race), a global optimum is found. However, if the unconnected path wins the race, Race$^{\text{con}}$ turns into a so-called deceptive function that gives hints to move away from all global optima and to get stuck in a local optimum. In this situation, the expected time to reach a global optimum is exponential, i.e., $2^{\Omega(n^\varepsilon)}$ for some constant $\varepsilon > 0$. For Race$^{\text{uncon}}$, the (1+1) MA gets trapped in the same way if the connected path wins and a global optimum is found in case the unconnected path wins.

The precise result is as follows. The preconditions $\delta \geq 36$, $\delta/\tau \geq 2/n$, and $\tau = O(n^3)$ require that "enough" iterations of local search are performed during a polynomial number of generations. The reason is that local search must be a visible component in the algorithm for the different local search frequencies to take effect. The condition $\tau = n^{\Omega(1)}$ as well as the choice of the initial search point are required for technical reasons.

**Theorem 4.** *Let $\delta = poly(n)$, $\delta \geq 36$, $\delta/\tau \geq 2/n$, $\tau = n^{\Omega(1)}$, and $\tau = O(n^3)$. If the (1+1) MA starts with a search point whose positions on the connected and unconnected paths are 0 and $n^5$, respectively, then with overwhelming probability*

- *the (1+1) MA with local search frequency $1/\tau$ optimizes Race$^{\text{con}}$ in polynomial time while the (1+1) MA with local search frequency $1/(2\tau)$ needs exponential time on Race$^{\text{con}}$ and*

- *the (1+1) MA with local search frequency $1/\tau$ needs exponential time on Race$^{\text{uncon}}$ while the (1+1) MA with local search frequency $1/(2\tau)$ optimizes Race$^{\text{uncon}}$ in polynomial time.*

The proof is quite technical; it requires good estimations for the progress made on the connected and the unconnected path, respectively. This is done separately for generations with and without local search, respectively. Using appropriate values for the lengths of the two paths derived from the analysis, one can show the following with overwhelming probability. With local search frequency $1/\tau$, within $n^4$ generations on both race functions the end of the connected path is reached first. On Race$^{\text{con}}$ the (1+1) MA has then found on optimum, while it has become trapped on Race$^{\text{uncon}}$. With local search frequency $1/(2\tau)$, within $\sqrt{2}n^4$ generations the total progress by local search on the connected path is decreased by a factor of roughly $1/\sqrt{2}$, compared to the previous setting. At the same time, the total progress on the unconnected path by mutation is increased by a factor of roughly $\sqrt{2}$. Summing up the progress values yields that then with overwhelming probability the (1+1) MA has found the end of the unconnected path first and Race$^{\text{uncon}}$ is optimized, while the (1+1) MA is trapped on Race$^{\text{con}}$.

An interesting insight gained from the analysis is that just one iteration of local search helps significantly with the location of isolated peaks. Mutation has to flip three specific bits in order to reach the next point on the unconnected path. However, if local search is called after mutation and were it only for one iteration, the next point on the unconnected path is also reached if only two out of the mentioned three bits are flipped. The probability for a successful step is hence by a factor of roughly $3n$ larger! So, in contrast to our intuition, local search does indeed help to optimize the unconnected path. Fortunately for our proof, the steps made on the unconnected path in generations with local search are unbiased. Creating the next successor on the unconnected path by mutation and local search has the same probability as creating the closest predecessor on the path. Both operations will be accepted with high probability if $\delta \geq 6$ since at least $\delta - 1$ iterations of local search are spent to make progress on the connected path. Also recall that the connected path is weighted with a factor of $n$ in the fitness function. Hence, the progress made on the connected path will dominate the effect of movements on the unconnected path. The search on the unconnected path in generations with local search is hence unbiased and the probability of making large progress due to the random walk behavior can be bounded. However, this only holds under the condition that $\tau = n^{\Omega(1)}$, i.e., if the local search frequency is not too high. Otherwise, the variance of the random walk behavior will indeed have a significant effect on the progress on the unconnected path. The author conjectures that with a very high local search frequency, the effect might even be reversed such that the unconnected path has a larger benefit from local search than the connected path.

## 5 Conclusions

Finding a good balance between global and local search is a crucial step in the design of memetic algorithms. This topic has been addressed explicitly or implicitly in a variety of applications as well as in empirical and theoretical works. An important conclusion is that the optimal balance is determined by many aspects. Finding a good balance involves knowledge on the problem structure as well as a careful consideration of the algorithms' operators, local search neighborhoods, settings of other parameters like the mutation strength, and implementation issues.

The chapter also covered theoretical approaches that are important to memetic algorithm researchers as they show the limits on the efficiency of memetic algorithms. The time complexity of local search is an interesting and rich topic in its own right and it can help to understand the effect of local search in memetic settings. PLS-completeness results indicate that for many practical problems there probably is no algorithm that can always find a local optimum in polynomial time. These intractability results can help prevent researchers from trying to achieve the impossible. Finally, running time analyses for a particular memetic algorithm have demonstrated superpolynomial or exponential performance gaps even for only slight changes of the parametrization. This rules out a priori design guidelines with polynomially related optimization times, for the considered memetic algorithm. All these findings indicate that finding a good parametrization remains an interesting and challenging topic for the years to come.

## Acknowledgments

The author was partially supported by a postdoctoral fellowship from the German Academic Exchange Service while visiting the International Computer Science Institute in Berkeley, CA, USA as well as by EPSRC grant EP/D052785/1.